\documentclass[transaction, letterpaper]{IEEEtran}
\ifCLASSINFOpdf
\else
\fi
\pdfoutput=1
\usepackage{amsmath}
\usepackage{etex}

\usepackage[utf8]{inputenc}
\usepackage[T1]{fontenc}

\usepackage[english]{babel}
\makeatletter
\adddialect\l@ENGLISH\l@english
\makeatother
\usepackage{csquotes}
\usepackage{authblk}
\usepackage{mathtools}
\usepackage{calc}
\usepackage{graphicx}
\usepackage{tikz}
\usepackage[backend=bibtex, url=false,%
			bibstyle=ieee,  
			citestyle=numeric-comp,     
			maxbibnames=3,%
			doi=false,isbn=false,eprint=false%
			]{biblatex}
\usepackage{array}
\usepackage{booktabs}
\usepackage{siunitx}

\usepackage{lmodern}
\usepackage{amsfonts}
\usepackage{amsmath}
\usepackage{textcomp}

\usepackage{bm}

\ifCLASSOPTIONcompsoc
\usepackage[caption=false,font=normalsize,labelfont=sf,textfont=sf]{subfig}
\else
\usepackage[caption=false,font=footnotesize]{subfig}
\fi

\newcommand{\figref}[1]{Fig.~\ref{#1}}  
\renewcommand{\vec}[1]{\ensuremath{\bm{#1}}}
\newcommand{\ind}[1]{\ensuremath{\left[ #1 \right]}}

\ifCLASSOPTIONcompsoc
\usepackage[caption=false,font=normalsize,labelfont=sf,textfont=sf]{subfig}
\else
\usepackage[caption=false,font=footnotesize]{subfig}
\fi

\begin{document}
%
\title{Biologically Inspired Radio Signal Feature Extraction \\ with Sparse Denoising Autoencoders}
%
%
%

\author{Benjamin~Migliori,~Riley~Zeller-Townson,~Daniel~Grady,~Daniel~Gebhardt
\thanks{B. Migliori~(corresponding author) and D. Gebhardt are with SPAWAR Systems Center Pacific, 53560 Hull Street, San Diego, CA 92152.
		email: benjamin.migliori@navy.mil}
\thanks{R. Zeller-Townson is with Georgia Institute of Technology, Dept. of Biomedical Engineering, North Ave NW, Atlanta, GA 30332}%
\thanks{D. Grady is with ID Analytics, 15253 Avenue of Science, San Diego, CA 92128}
\thanks{Submitted for online publication, Sept 25, 2015. Revised Feb 29, 2016.}
}

\maketitle
\begin{abstract}
Automatic modulation classification (AMC) is an important task for modern communication systems; however, it is a challenging problem when signal features and precise models for generating each modulation may be unknown.  We present a new biologically-inspired AMC method without the need for models or manually specified features --- thus removing the requirement for expert prior knowledge.  We accomplish this task using regularized stacked sparse denoising autoencoders (SSDAs).  Our method selects efficient classification features directly from raw in-phase/quadrature (I/Q) radio signals in an unsupervised manner.  These features are then used to construct higher-complexity abstract features which can be used for automatic modulation classification.  We demonstrate this process using a dataset generated with a software defined radio, consisting of random input bits encoded in 100-sample segments of various common digital radio modulations. Our results show correct classification rates of > 99\% at 7.5 dB signal-to-noise ratio (SNR) and > 92\% at 0 dB SNR in a 6-way classification test. Our experiments demonstrate a dramatically new and broadly applicable mechanism for performing AMC and related tasks without the need for expert-defined or modulation-specific signal information.
\end{abstract}


%
\IEEEpeerreviewmaketitle

\section{Introduction}
\label{introduction}

\begin{figure}
	\centering
	\subfloat[ Likelihood-based modulation classification\label{fig:amc-methodologies-lb}, requiring \textit{a priori} knowledge of the probability distribution function of the received waveform conditioned on the details of the modulation. ]{ \includegraphics[trim=0.in 5.2in  6.5in 0.2in,clip, width=\linewidth, ]{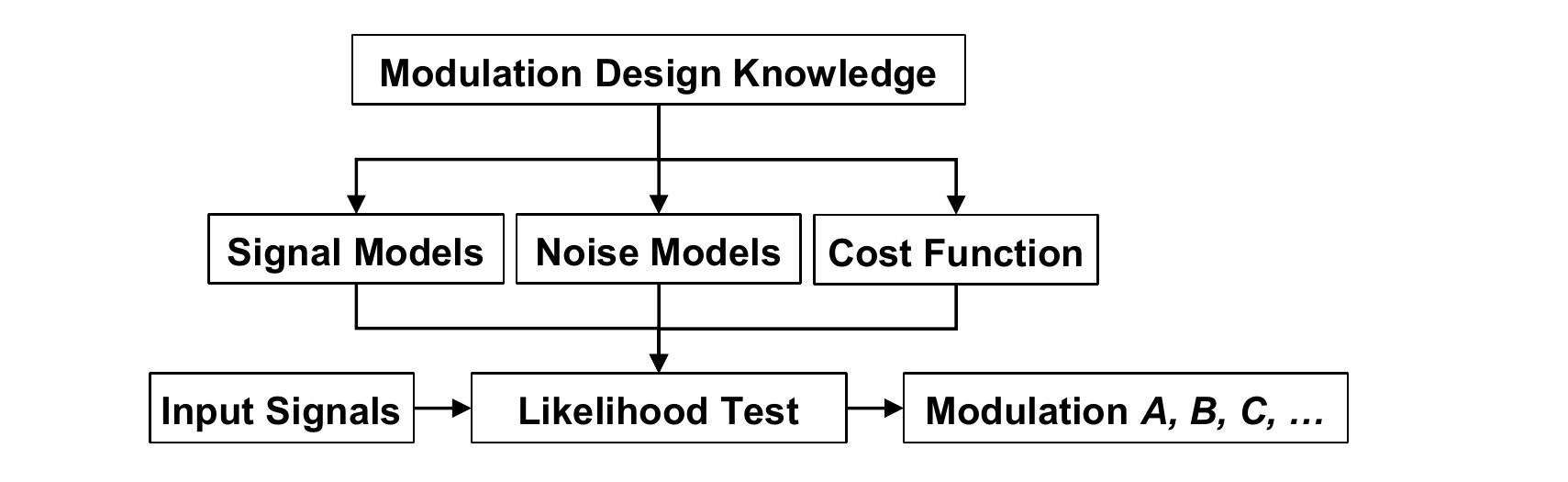} }\hfill
	\subfloat[ Feature-based modulation classification\label{fig:amc-methodologies-fb}, requiring known characteristics of each modulation to be classified. ]{ \includegraphics[trim=0.in 2.8in  6.5in 2.7in,clip, width=\linewidth, ]{migliori-fig1.pdf} }\hfill
	\subfloat[ Biologically-inspired automatic modulation classification\label{fig:amc-methodologies-sdae}, with the only requirement being a set of observed signals with labeled modulations.]{ \includegraphics[trim=0.in 0.5in  6.5in 5.4in,clip, width=\linewidth, ]{migliori-fig1.pdf}} %
	\caption[]{ Comparison of workflows for automatic modulation classification. Existing methodologies shown in \subref{fig:amc-methodologies-lb} and \subref{fig:amc-methodologies-fb}, and ours in \subref{fig:amc-methodologies-sdae}. Note that \subref{fig:amc-methodologies-sdae} does not require expert or situational knowledge, and instead only uses observational data. }
	\label{fig:amc-methodologies}
\end{figure}

\IEEEPARstart{B}{lind} identification of signal modulations is a difficult task that bridges signal detection and the creation of useful information from received signals. This task is even more challenging in a non-cooperative or noisy environment with realistic channel properties, even with prior knowledge of the modulations to be detected. When such information is not available, classification is generally not feasible as most existing methods require prior information regarding the modulation mechanism. Broadly, automatic modulation classification (AMC) techniques fall into two categories: likelihood-based (LB) and feature-based (FB)~\cite{Dobre2007}. In LB AMC, the likelihood function of the received signal belonging to a modulation is used to create a likelihood ratio, which is compared to a pre-determined decision threshold~\cite{Polydoros1990,xu2011likelihood}. LB AMC is optimal from a theoretical Bayesian perspective, in that it minimizes the chance of a wrong classification. However, it often has high computational complexity and requires careful design and selection of signal and noise models.  Feature-based (FB) AMC uses expert-selected or designed signal filters based on known characteristics of expected modulations in a decision tree with associated thresholds to determine the detected modulation family~\cite{Soliman1992,Ramkumar2009}. LB (\figref{fig:amc-methodologies-lb}) and FB (\figref{fig:amc-methodologies-fb}) methods both require substantial design-side knowledge about the modulation properties, and make specific assumptions regarding environmental noise. 

Contrast an AMC task with that of an animal moving within a natural environment. Animal sensory systems, such as vision and audition, have evolved over millions of years to detect, identify, and respond to novel events that could pose a threat or indicate a reward. As a result, when a new sound or sight is observed, most animals will make an immediate decision to classify it as friend, foe, or neutral; they perform this task without an explicit model or expert knowledge of the environment. Instead, they rely on previously learned low-level environmental features (such as edges and luminance transitions in vision) that generate activity in the different layers of neurons within sensory cortex~\cite{pmid14449617}. As the information propagates through layers of cortex, the concepts that the neurons are sensitive to become more and more abstract~\cite{pmid15056711,hegde2000selectivity}. Decisions based on these hierarchical features (called receptive fields) are what allow the animal to make the friend-foe decision~\cite{Endler215}. This decision can be made without having prior knowledge of the exact input properties, and in the presence of noise or corruption; further, the process is naturally suited to non-cooperative environments.

The receptive fields of organisms often possess striking features that are not explained by simple statistical approaches. In many mammalian visual systems, the receptive fields are spatially localized, oriented, and scale-dependent Gabor-like filters; these features are not generally recovered by searching for orthogonal basis functions in the space of natural images. It appears that one important criterion of biological receptive fields is to maximize the statistical independence of the basis functions~\cite{Olshausen1996, Bell1997}; one method for accomplishing this is to derive a sparse over-complete basis, such that any particular basis function is highly selective to a small number of environmental features~\cite{Olshausen1996, Olshausen2004, NIPS2011_1115}. In visual systems, this manifests in the specificity of neural populations to particular stimuli. It may also be indicative of an efficient implementation in terms of energy use and reaction time, as compared to a more uniformly distributed representation (an important consideration for evolution)~\cite{Endler215}.

Here, we demonstrate that a biologically-inspired artificial neural network, configured such that it recreates Gabor-like receptive fields (\figref{fig:gabor}) when trained on natural images, can generate useful information regarding a non-biological sensory input --- in this case, in-phase and quadrature (I/Q) signals acquired in the radio frequency spectrum. Our architecture ingests whitened I/Q signals (i.e. sensory input), uses stacked sparse denoising autoencoders (SSDAs) to adaptively generate primitive and complex features (i.e. receptive fields), and then uses those features to perform automatic modulation classification. It requires as few as 100 I/Q timepoints to make a classification after training, and does so rapidly (microseconds/decision on a consumer laptop). The output of the SSDA is a high-dimensional representation that responds uniquely to signals of each modulation type. The system then translates this output to human-readable modulation labels using a training set (consisting of observed signals for each modulation) and a softmax multi-layer perceptron classifier. 

Our approach differs significantly from current approaches to automatic modulation classification in that it is model- and expert-free on the ingest side, and requires only example signals to detect future signals modulated in the same manner. 
This stands in sharp contrast to LB and FB methods, which require expert input at each stage. Each of these methodologies is compared in \figref{fig:amc-methodologies}. We have tested our methods on a variety of modulations (generated with a software-defined radio) and in additive white gaussian noise channels (AWGN). Our biologically-inspired algorithm demonstrates performance comparable with existing methods, yielding $> 99\% $ correct classification at 7.5~dB signal-to-noise ratio (SNR), and suggests a dramatically different approach for successful AMC in challenging environments or scenarios. 

Although modulated radio signals are in many ways quite different from the `signals' that biological neural systems have evolved to detect and process, they nonetheless propagate through an environment that produces many biologically relevant sources of noise, and it is reasonable to ask whether principles of biological sensing may lead to useful alternatives to existing statistical approaches for radio signal processing. As we demonstrate, our architecture does generate useful receptive fields that allow classification methods to better discriminate amongst classes; because these receptive fields arise from an unsupervised sparse encoding of the input, they represent the equivalent of useful primitive features within the space of man-made radio signals (much as edges and angles are primitives of visual spaces). This unsupervised feature selection and pre-training, which differentiates us from other neural-network based AMC methods~\cite{Wong2004, Aslam2012, Nandi1997}, makes our method generally applicable to a broad range of input modalities, potentially including images, network data flows, and other time-varying processes.

\section{Methods}
\label{methods}

\subsection{Problem Formulation}

Consider a (deterministic) system $M$ which accepts a signal $s$ and emits a prediction $M(s)$ of which one of a fixed number $N_{mod}$ of modulation families encodes $s$. Letting $T(s)$ be the true modulation of $s$, one way to characterize $M$ is with its full joint distribution, where $P_{c}^{(i'|i)}$ represents the probability of correct prediction for a given signal:

\begin{equation}
	\label{eq:joint-distribution}
	P_c^{(i'|i)} = \operatorname{Prob}\left[ M(s) = i' \text{ and } T(s) = i \right]
\end{equation}

A measure of the performance of $M$, and the one in which we are mainly interested, is the average correct classification across all tested modulation families, $P_{cc}$:

\begin{equation}\label{eq:pcc}
P_{cc} = \frac{1}{N_{mod}}\sum_{i=1}^{N_{mod}}P_c^{(i|i)}
\end{equation}
We also consider cases in which we allow $P_{cc}$ to be a function of signal to noise ratio (SNR).

\subsection{Radio Data Generation}
The data used for experimentation was synthetically generated radio signals, transmitted and received, but clean of outside interference.
We used the GNU Radio~\cite{GNURadio} software-defined radio (SDR) framework to construct the modulations that generated this data.

A binary file, produced by randomly choosing byte values $[0,255]$, is the waveforms' input.
This binary data is modulated as in-phase and quadrature I/Q samples using each of six methods: on-off keying (OOK), Gaussian frequency-shift keying (GFSK), Gaussian minimum-shift keying (GMSK), differential binary phase-shift keying (DBPSK), differential quadrature phase-shift keying (DQPSK), and orthogonal frequency-division multiplexing (OFDM).

For each modulation, the samples are sent to a BladeRF SDR, where they are upconverted to the carrier frequency. 
The SDR is configured in RF loop-back mode, such that the RF signal is sent and received only within the device's circuitry, and not to an external antenna. 
This arrangement provides added realism by incorporating the upconversion and radio effects, but without unwanted third-party signals that could pollute the controlled testing. 

The signal sampling rate is set so that the number of samples per symbol ($N_{SpS}$) is consistent for every modulation type, except for OFDM. In contrast with the other modulation techniques, OFDM encodes data on multiple carrier frequencies simultaneously, within the same symbol, and modulates each carrier frequency independently. Our experiment used an existing OFDM signal processing component that operates with a symbol rate different than the other configurations, but with the same sample rate.
This rate is identical for both the transmission and reception of the signal.
The received RF signal is down-converted at the radio and the resulting I/Q samples are stored for analysis.
\subsection{Training Data Generation and Preprocessing}

The data files need to be arranged into a format and structure for use by our neural network.
The I/Q data are split into segments consisting of $N_{SpV}$ samples, or samples per vector.
A segment is composed of interleaved I and Q values for each sample, forming a vector of length $2 \times N_{SpV}$.
Thus, each vector contains $ N_{SpV} \over N_{SpS} $ symbols.
These vectors are placed into two sets,  \emph{train} and \emph{test} (sizes $N_{Vtrain}$ and $N_{Vtest}$), such that both the modulation type and positions within the set are random.
The parameter $N_{SpV}$ is identical for each modulation type for all experiments described in this paper.
The specific values of all parameters are shown in Table~\ref{table_params}.

\begin{table}[htp]
\caption{Data generation parameters}
\begin{center}
\begin{tabular}{ l  l  l}
\toprule
\textbf{Description}  &  \textbf{Parameter}  &  \textbf{Value}\\
\midrule
samples per symbol  &  $N_{SpS}$ &  10\\
samples per vector  &  $N_{SpV}$ &  100\\
number of training vectors  &  $N_{Vtrain}$  &  60000\\
number of training vectors per modulation & $N_{Vmod}$ & 10000 \\
number of test vectors &  $N_{Vtest}$  &  10000\\
\bottomrule
\end{tabular}
\end{center}
\label{table_params}
\end{table}%

We use a pre-processing step which consists of fitting a ZCA whitening filter $Z$~\cite{Bell1997} to the training set.
\subsection{Neural Network Architecture}
Our neural network consists of a series of pre-trained sparse denoising autoencoders (SDAs)~\cite{Vincent2008, Olshausen2004}, followed by a fully-connected softmax classifier layer.  Starting from a signal sample vector $\vec{s}$ as described in the previous section, we compute $\vec{x} = Z \cdot \vec{s}$. The input units of the first autoencoder are set to the values given by $\vec{x}$; the values of the hidden layer units are calculated according to 
\begin{equation}
\vec{y} = \sigma(W \cdot c(\vec{x}) + \vec{b}_v) 
\end{equation}
and output layer values are calculated as 
\begin{equation}
\vec{z} = \sigma(W^T \cdot \vec{y} + \vec{b}_h) 
\end{equation} 
Here, $\sigma$ is a non-linear activation function that operates element-wise on its argument, and $c$ is a stochastic ``corruptor'' which adds noise according to some noise model to its input. $c$ is non-deterministic: $c$ may corrupt the same sample vector $\vec{x}$ in different ways every time $\vec{x}$ is passed through it.  After training, the output layer of each autoencoder is discarded, and the hidden layer activations are used as the input layer to the next autoencoder. We hypothesized that an overly sparse or compact representation would be unable to distinguish between identical modulations shifted in time. Thus, the number of neurons on the first and second layers were chosen such that with fully sparse activation constraints (5\% of total neurons), there would still be a significant number of neurons active for a given sample (i.e. $\sim25$).    

We explored seven total architectures, summarized in \autoref{table_archs}. These included a simple softmax classifier, a two-layer MLP without pre-training, two one-layer denoising autoencoders (with and without sparsity or L2 regularization), two two-layer autoencoders (with and without sparsity or L2 regularization), and a deep SSDA with regularization (5 layers). The exact number of neurons (500 in layer 1 and 2, 250 in layers 3 and 4, and 100 in layer 5) was chosen arbitrarily to conform to available computing resources. To prevent learning of a trivial mapping, either the layer-to layer dimensionality or sparsity constraint was altered between each pair of layers.  

The parameters of a single autoencoder are the weight matrix $W$ and bias vectors $\vec{b}_v$ and $\vec{b}_h$; unsupervised training is the process of adjusting these parameters so that the output layer reproduces the input as precisely as possible while also subjecting it to a constraint designed to encourage ``sparse activation'' of hidden layer units, that is, to encourage hidden layer unit activations to remain near $0$ except for a small fraction. The overall cost function for a single autoencoder layer is

\begin{equation} 
J(W, \vec{b}_v, \vec{b}_h) = \left< \left\lVert \vec{z}_i - \vec{x}_i \right\rVert^2 \right>_i + \beta \sum_k \operatorname{KL}(\rho, \rho_k) 
\end{equation}

Here, $i$ indexes over data vectors and $k$ indexes over hidden layer units. $\beta$ and $\rho$ are parameters, $\vec{x}_i$ is the $i$-th data vector, $\vec{z}_i$ is the corresponding output layer activation, $\rho_k$ is the average activation level of the $k$-th hidden unit over all data vectors, and $\operatorname{KL}(\rho, \rho_k) = \rho \log \frac{\rho}{\rho_k} + (1-\rho) \log \frac{1-\rho}{1-\rho_k}$ is the Kullback-Leibler divergence.

The hidden layer activations of one autoencoder can be supplied as the input to another autoencoder, leading to a stacked architecture. Denote the input, hidden, and output units of a single autoencoder at layer $\ell$ as $\vec{x}^{(\ell)}$, $\vec{y}^{(\ell)}$, $\vec{z}^{(\ell)}$ respectively. Then the process of forward propagation through the entire network of autoencoders proceeds sequentially according to
\begin{equation}\label{eq:forward-propagation}
\vec{y}^{(\ell)} = \sigma\left( W^{(\ell)} \cdot c_\ell(\vec{y}^{(\ell-1)}) + \vec{b}_v^{(\ell)} \right)
\end{equation}
for $\ell = 1 \dots L$, and with the convention that $\vec{y}^{(0)}$ is the input layer.

We conduct sequential, unsupervised training of individual autoencoder layers using stochastic gradient descent with a batch size of $100$ and the AdaGrad method~\cite{Duchi2011} based on the I/Q data set described previously. The parameters used for training are listed in Table \ref{tab:fixed-training-params}.

\begin{table}\centering
	\caption{\label{tab:fixed-training-params}Parameters used for Training.}
	\begin{center}
	\begin{tabular}{l l l}
		\toprule
		\textbf{Description}  &  \textbf{Symbol}  &  \textbf{Value}\\
		\midrule
	activation function & $\sigma$ & $\tanh$ \\
	layer 1 corruption &$c_1$ & Bernoulli, $p_{flip} = 0.2$ \\
	layer 2 corruption &$c_2$ & Bernoulli, $p_{flip} = 0.3$ \\	
	layer 1 sparsity target &$\rho_1$ & $0.05$ \\
	layer 2 sparsity target &$\rho_2$ & $0.00$ \\
	\bottomrule
	\end{tabular}
	\end{center}
\end{table}

\subsection{Supervised fine-tuning and classification}

We follow the unsupervised pre-training phase with supervised fine-tuning. For this phase we organize the pre-trained autoencoders into a purely feed-forward multilayer perceptron according to \autoref{eq:forward-propagation}, with an additional final layer 
\begin{equation}
\vec{y}^{(L)} = \operatorname{softmax} \left( W^{(L)} \cdot \vec{y}^{(L-1)} + \vec{b}^{(L)} \right)
\end{equation}

\begin{figure*}
	\centering
	\includegraphics[page=1,trim=0.2in 5.8in  3in 0in,clip, width=0.7\linewidth, ]{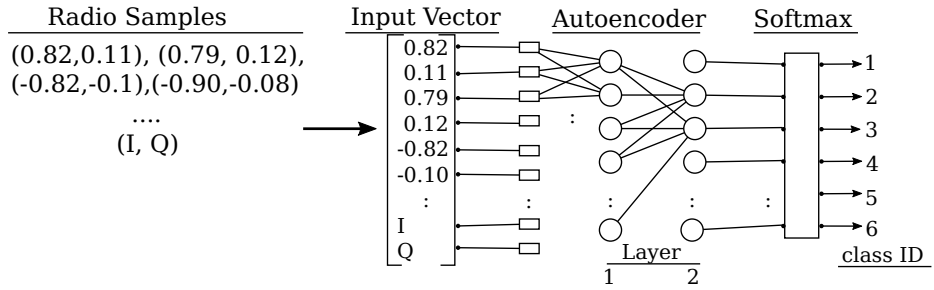}
	\caption{Architecture of input data, neuron organization, and output classification.  The autoencoders are configured with sparsity across neurons and noise corruption between layers. The softmax is configured to use a regularization constraint during refinement of the autoencoder weights.}
	\label{fig:architecture}
\end{figure*}

Interpreting the final output vector of the multilayer perceptron as a probability distribution over modulation families, supervised learning attempts to minimize the negative log-likelihood function with an additional L2 regularization term to encourage the model to retain the sparsely activating features learned during the unsupervised phase. The regularization term $\lambda$ was set to a value of 1 or 0, depending on the desired experiment configuration. Explicitly, where $n$ is the list of samples, $L$ is the total number of layers, $\vec{y^{(\ell)}}$ is the output of layer $l$, and $W^{\ell}$ indicates the weight matrix between layers $l$ and $l+1$,  the loss function of the multi-layer perceptron is given by:
\begin{equation}\label{eq:loss}
J = -\frac{1}{n}\sum_{i=1}^n \left( \log{\frac{e^{y_{t_i}^{(L)}}}{\sum_{m=1}^{s_L} e^{y_{m_i}^{(L)}}}} \right ) +\lambda\sum_{\ell=0}^{L-1} \sum_{k=1}^{s_\ell}\sum_{j=1}^{s_{\ell+1}}\left(W_{jk}^{(\ell)} \right )^2
\end{equation}
$t_i$ indicates the index corresponding to the correct label for sample $i$, and $s_\ell$ is the number of units in layer $\ell$.  We minimize \autoref{eq:loss} using batch stochastic gradient descent \cite{bottou2010large}.  
The overall architecture is shown in \figref{fig:architecture}. We test several variations of the stacked autoencoder architecture, as described below, but the parameters which we hold fixed in all our experiments are listed in \autoref{tab:fixed-training-params}.

\subsection{Addition of noise}

\begin{figure*}
	\centering
	\subfloat[20 dB signal-to-noise.]{%
		\includegraphics[ trim={0 280 0 0px},clip,width=\linewidth]{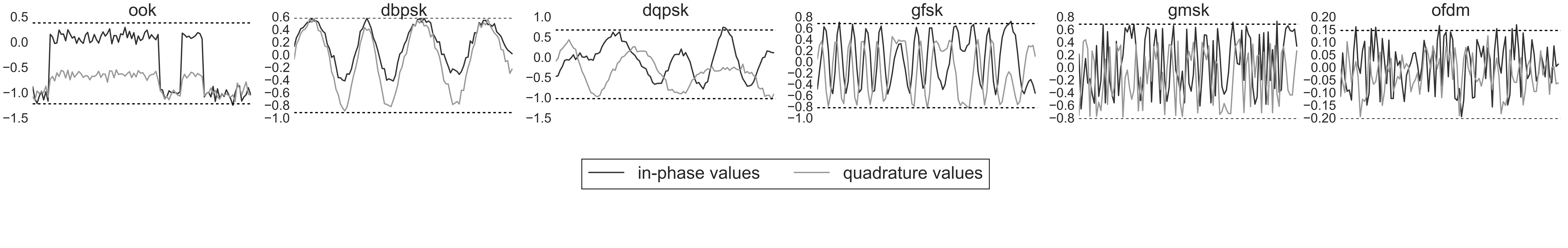}}\hfill
	\subfloat[0 dB signal-to-noise.]{%
		\includegraphics[ trim={0 280px 0 50px},clip,width=\linewidth]{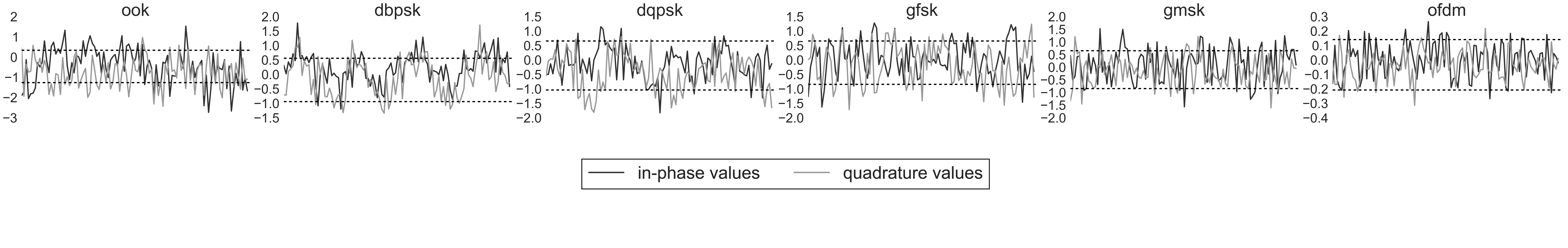}}\hfill
	\subfloat[-6 dB signal-to-noise.]{%
		\includegraphics[ trim={0 280px 0 50px},clip,width=\linewidth]{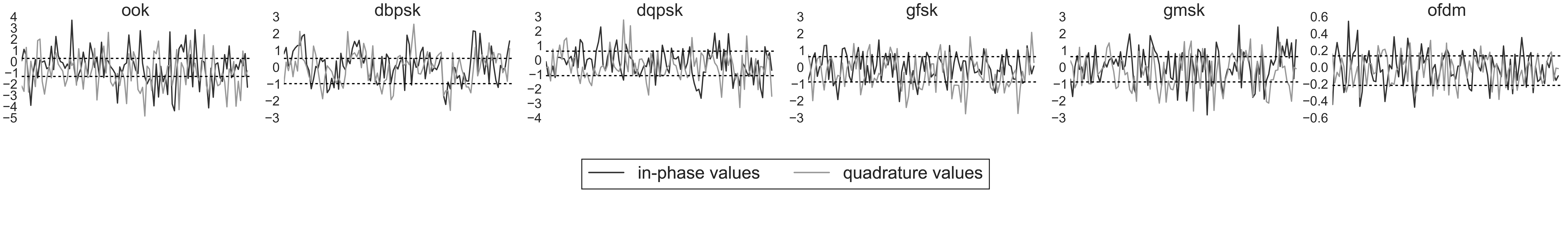}}\hfill	
	\subfloat{%
		\includegraphics[ trim={50px 100px 100px 320px},clip,width=\linewidth]{migliori-fig3c.png}}\hfill	
	\caption{Example input vectors for each modulation type under varying noise. Each instance has 100 (I,Q) samples. Dashed lines indicate the envelope of the 20 dB SNR signals.}
	\label{fig:data_modulations}
\end{figure*}

To assess the performance of our system with a more realistic channel model, we altered the test data set with additive white Gaussian noise (AWGN).
These data configurations were used as input in a purely feed-forward mode, in that our system was not re-trained, and its modulation classification output evaluated. AWGN was added to each set of signal modulation types such that for each set the resulting signal-to-noise ratio (SNR) matches a given value.
This was necessary since each modulation type, as sampled by the radio, had different average power levels.
For each of these signal modulation sets, $\{S_{mod}\}$, the added noise power, $P_{noise}$ is:
\begin{equation}
P_{noise} = \beta \cdot \frac{1}{N_{s(mod)}}\sum_{\{S_{mod}\}}\frac{1}{\tau}\sum_{t=1}^\tau [s_t]^2 
\end{equation} 
 where $N_{s(mod)}$ is the number of sample vectors for a particular modulation, $s_t$ is an individual signal sample vector of length $\tau$, and $\beta$ is a factor chosen such that $10\log(P_{\{S\}}/P_{noise})$ matches the desired SNR.
 Examples of modulation data with the addition of noise are shown in~\figref{fig:data_modulations}.  Note that all modulations exhibited similar transmitted power with the exception of OOK, which was slightly larger.

\section{Results}
\label{results}

\subsection{Classifier convergence and accuracy}
We measured the overall classification accuracy $P_{cc}$ (Equation \ref{eq:pcc}) for each architecture.  These architectures varied in the number of layers and the types of cost enforced during training; we used a cost for non-sparse activation as an L1 penalty (sparsity), and a cost for weight magnitude as an L2 penalty (weight decay). The architectures included a softmax perceptron (without sparsity or weight decay costs), a single layer sparse denoising autoencoder with and without weight decay costs, and a two-layer sparse denoising autoencoder with and without layer 2 sparsity costs and weight decay costs. We also tested a deep SSDA with five layers.  The architectures were chosen to study the effects of adding additional regularizations on the ability of the system to classify radio modulations. The results for each architecture are summarized in Table \ref{table_archs}. Architectures A, C, D, and E performed approximately two orders of magnitude better than the softmax classifier alone on the test set in the absence of noise.  With both L2 regularization and sparsity constraints, the number of training examples required to obtain convergence increased, and in particular architecture D required significantly more time to converge than the others. However, this is offset by the increased performance in the presence of channel noise.

\begin{table*}\centering
	\caption{\label{table_archs}Experimental Architectures and Misclassification Rates}
		\begin{tabular}{ c S S c c c }
			\toprule
			Label & {$P_{cc}$ (\%)} & {$P_{cc}$ (\SI{0}{\decibel}) (\%)} & Neurons ($N_1$/$N_2$) & Sparsity ($\rho_1$/$\rho_2$) & Regularization \\
			\midrule
			Softmax Only & 46.9 &  36.6 & ---/--- &  ---/---   & N/A\\
			MLP	Only	 & 55.6 &    & ---/--- &  ---/---   & Yes, Dropout\\
			A        & 99.91 &  64.9 & 500/--- &  0.05/---  & No\\
			B        & 90.8 &  73.0 & 500/--- &  0.05/---  & Yes\\
			C        & 99.86 &  74.7 & 500/500 &  0.05/---  & No\\
			D        & 99.56 &  91.9 & 500/500 &  0.05/0.00 & Yes\\ 
			E		 & 99.10 &  65.0 & 500/500/250/250/100 & 0.05/0.00/0.10/0.00/0.25 & Yes\\
			\bottomrule
		\end{tabular}
\end{table*}

\subsection{Classifier performance in the presence of channel noise}

\begin{figure*}
	\centering
	\includegraphics[trim=2.25in 3.5in  3.in 1.0in,clip, width=\linewidth, ]{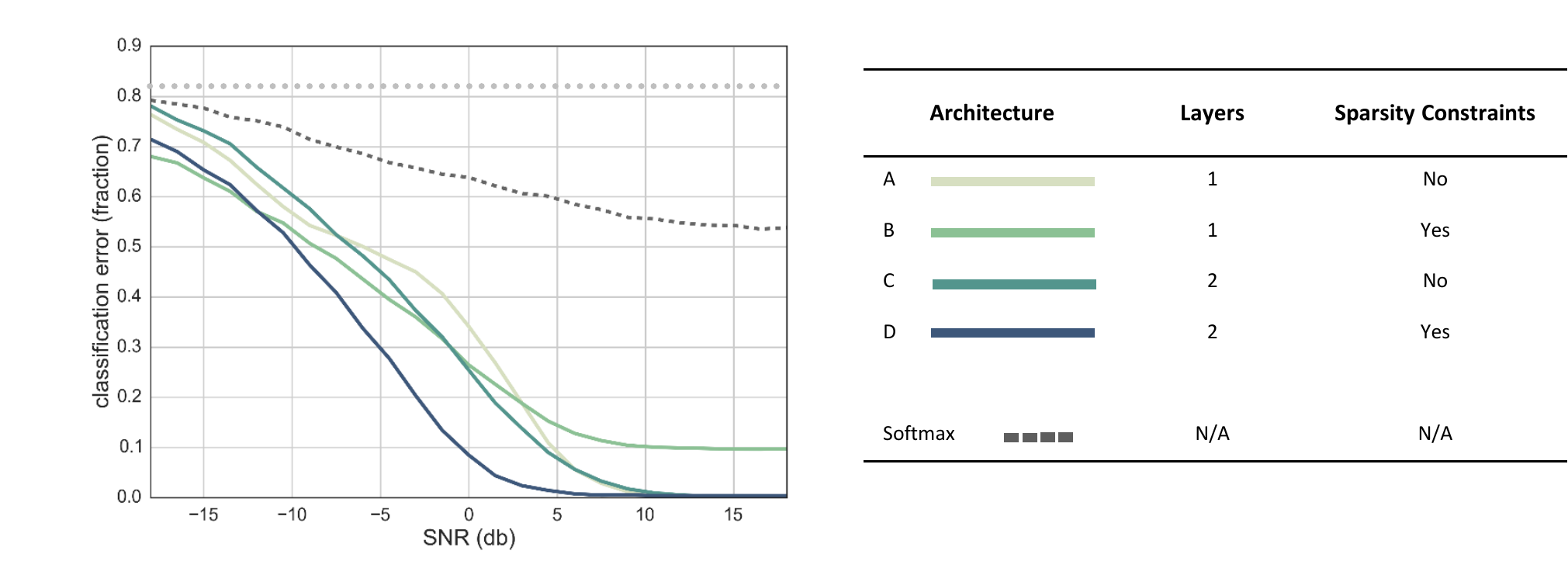}
	\caption{\label{fig:snr} Classification error ($1-P_{cc}$) on the test set as a function of noise. For each example in the test set, we add Gaussian noise to produce the desired signal-to-noise ratio before presenting the example to the neural network. The configurations correspond to Table \ref{table_archs}.  The classification error of a random guess is indicated by the horizontal dotted line. A value of 0.00 corresponds to perfect classification.  Configuration E had substantially worse performance and is not included in this plot.}
\end{figure*}

The ability to classify modulations under low signal-to-noise ratios (SNR) is one of the crucial abilities of a successful AMC algorithm. We tested our system's performance by measuring $P_{cc}$ as a function of SNR. Our AMC method degrades gracefully as the SNR decreases, and approaches random chance ($P_{cc}=1/6$) at $\approx \SI{-20}{\dB}$.  The performance of each configuration is shown in \figref{fig:snr} and summarized in Table \ref{table_archs}.

In the absence of weight decay costs (L2 penalties), a single-layer sparse denoising autoencoder performs better than a softmax classifier across the SNR range studied (A, \figref{fig:snr}). A single-layer SDA with weight decay (B, \figref{fig:snr}) substantially improved the generalization of our classifier at higher noise levels (as expected from \cite{moody1995simple}), but prevented the network from converging to a reasonable level of accuracy (i.e. $P_{cc}>>5/6$).  To allow a better fit of the data to occur, we added a second layer (without weight decay or sparsity costs, C in \figref{fig:snr}).  This improved the overall performance. However, a two-layer sparse stacked denoising autoencoder with weight decay and sparsity costs in layer 2 (D, \figref{fig:snr}) performs significantly better across all SNRs, with an error rate at 0 dB SNR of 8\% and a performance > 5 dB better than the closest competitor (C in \figref{fig:snr}).  These results indicate that sparsity, multiple layers, and regularization during further training are important to achieve generalization of the classifier. 

For applications to real signals, the magnitude of the integral under the curve in \figref{fig:snr} is somewhat more important than maximal classification accuracy, so we consider architecture D for further discussion. 

\subsection{Confusion matrix and per-type classification accuracy}

\begin{figure*}
	\centering
	\includegraphics[width=6in]{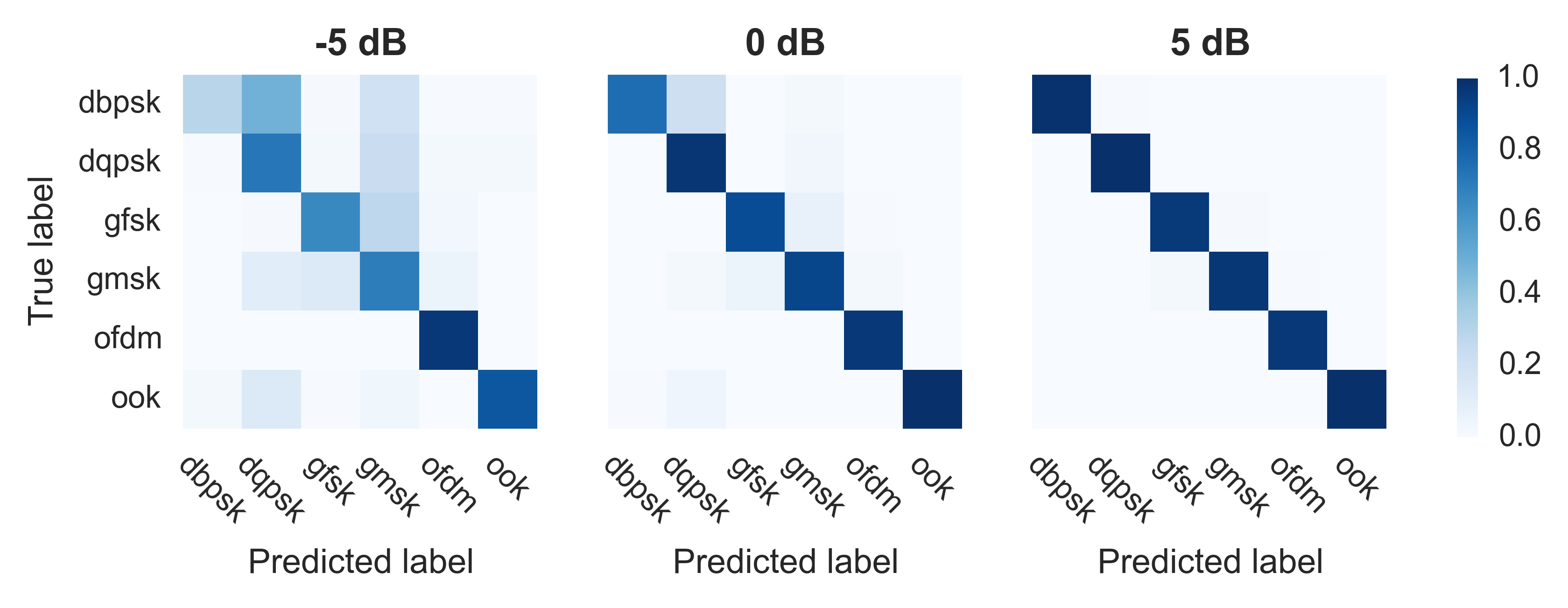}
	\caption{\label{fig:confusion-matrix} Confusion matrix for each modulation family as classified by architecture D (see Table \ref{table_archs}) at a signal-to-noise ratio of \SI{-5}{\decibel}, \SI{0}{\decibel}, and \SI{5}{\decibel}. These are visualizations of the empirically determined joint distribution of~\autoref{eq:joint-distribution}.}
\end{figure*}

Although $P_{cc}$ is a good indication of classifier performance overall, we are also interested in identifying specific modulations that may be more or less challenging for our method.  To do this, we construct a confusion matrix of dimension $N_{mod} \times N_{mod}$ consisting of the values of $P_c^{(i'|i)}$. We plot the confusion matrix for the classifier with the highest overall performance in \figref{fig:confusion-matrix} at SNRs of \SI{-5}{\dB}, \SI{0}{\dB}, and \SI{5}{\dB}. Signals that use on-off keying are the easiest to classify, and virtually none are misclassified at these noise levels. Of the remaining modulation families, there is some error in all of them as SNR decreases; the classifier tends to over-predict GMSK at the expense of other types. Another error is the  confusion of the DQPSK and DBPSK modes at high noise.  

\begin{figure}
	\centering
	\includegraphics[width=3in]{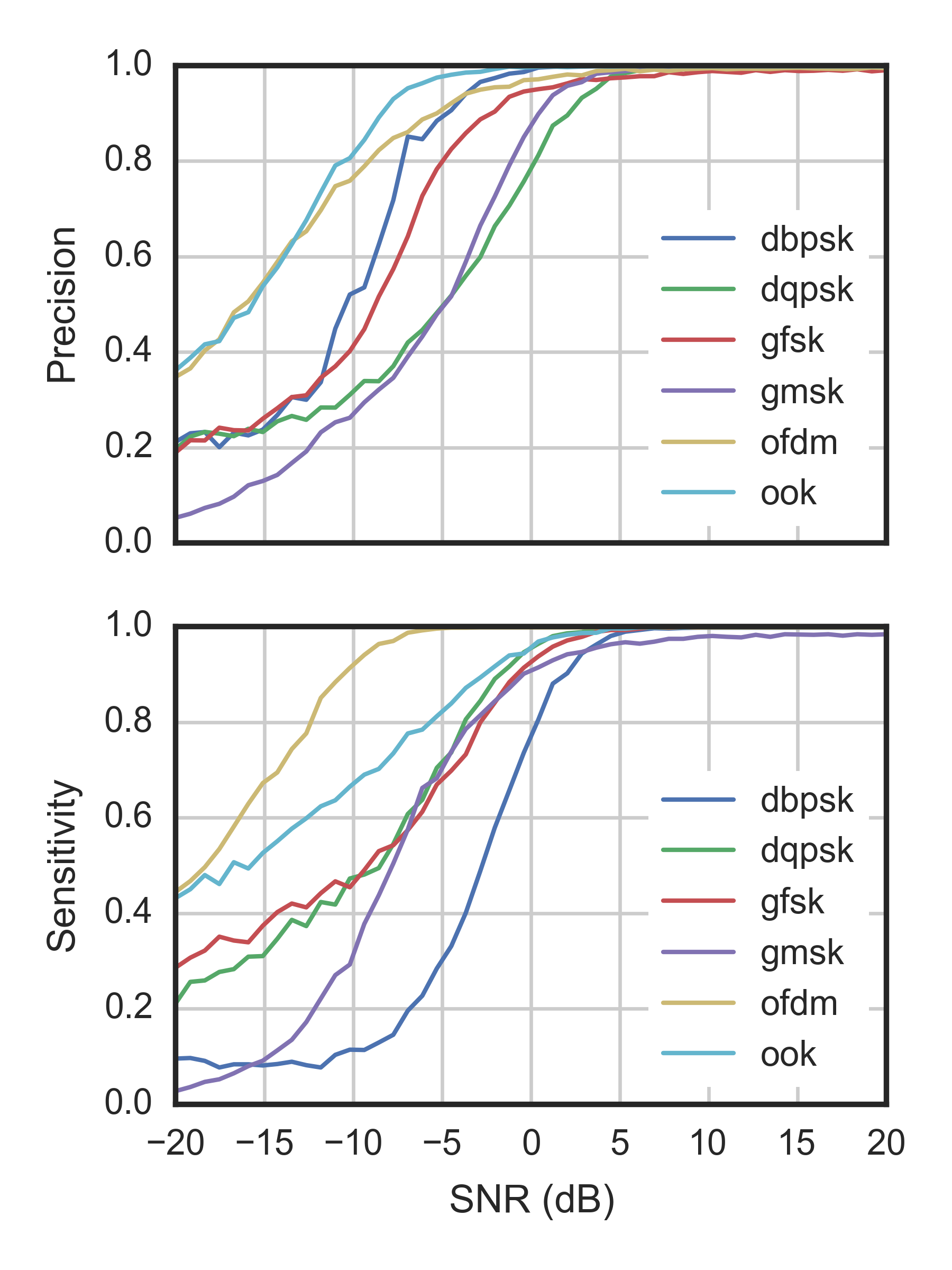}
	\caption{\label{fig:precision}Precision (\autoref{eq:precision}) and sensitivity (\autoref{eq:sensitivity}) curves for each modulation family for configuration D (see Table \ref{table_archs}) as a function of noise.}
\end{figure}

Parallel to this, we also examine the precision and sensitivity of the classifier to each modulation family as a function of the SNR{}. Let $m_i$ and $y_i$ be the true and predicted class label, respectively, for sample $i$. Then the precision of the classifier for class $k$ is
\begin{equation}\label{eq:precision}
P_k = \frac{\sum \ind{m_i = k \text{ and } y_i = k}}{\sum \ind{y_i = k}}
\end{equation}
and its sensitivity is
\begin{equation}
\label{eq:sensitivity}
S_k = \frac{\sum \ind{m_i = k \text{ and } y_i = k}}{\sum \ind{m_i = k}}
\end{equation}
where brackets are the indicator function ($\ind{p}=1$ if $p$ is true and $0$ otherwise). We plot the precision and sensitivity of the highest performance classifier (D, \autoref{table_archs}) in \figref{fig:precision}. These results confirm that on-off keying is extremely robust to noise under this classification system, and that precision for each modulation family falls off as noise increases. However, sensitivity (the number of correctly identified signals from each modulation total) varies much more strongly across the different modulation families. In particular, at -10 dB SNR we observe sensitivities ranging from 0.1 to 0.9 depending on the modulation type. 

\section{Discussion}
\label{discussion}
\subsection{Comparison to Traditional Modulation Classification}
Performance characterization and comparison of our system with more traditional methods is challenging, as modulation classification is a large and diverse field of study. However, a recent review paper on automatic modulation classification~\cite{Dobre2007} makes a valiant effort at summarizing and comparing likelihood-based (LB) and feature-based (FB) classifications on the same footing. In this review, in a 2-way classification task between BPSK and QPSK modulations with an average-likelihood ratio test method (ALRT), $P_{cc}$ of 97.5 \% at \SI{-3}{\dB} was observed. While this performance exceeds that of our method, ALRT classification is strongly tied to modulation type; for a QAM discrimination task, an SNR of \SI{7}{\dB} is required to obtain $P_{cc}=99 \%$ for the same classifier. Worse, in cases where model parameters such as exact carrier phase are unknown, performance begins to drop off; SNRs of  \SI{30}{\dB} are required to achieve $P_{cc}=88\%$ with a quasi-ALRT classifier choosing between 16QAM and V.29 modulations. Similarly, FB methods perform well at high SNRs (typically with $P_{cc} > \approx 95\%$ at SNRs $>$ \SI{5}{\dB}) but suffer from significant performance variations for different modulation types. 

The performance of our system ($P_{cc} =92\%$ at \SI{0}{\dB} in a 6-way AMC task) is competitive when compared with the performance of AMC using LB or FB methods, as well as ANN-based FB methods~\cite{Wong2004}. Crucially, unlike existing methods, prior knowledge of modulation design or characteristics is completely unnecessary for our method; to duplicate the performance we have observed, one must merely obtain a list of signals that are known to be different from each other. It is not unreasonable to expect that crude methods of signal discrimination could be used to build such a training dataset with environmentally-corrupted, non-cooperative, or entirely novel signals, which would then allow a method such as ours to perform modulation classification where the other methods discussed would be entirely unfeasible. Additionally, our method was evaluated on sequences of 10 symbols or 100 I/Q timepoints; this is substantially fewer timepoints than most existing AMC methods use, and makes our system more likely to be valuable for classification in dynamically shifting environments. 

\subsection{Pre-training is crucial to AMC performance}
Our results indicate that the use of unsupervised pre-training is crucial to the AMC task. We observed this by exploring the overall classification performance of our SSDA network vs a multi-layer perceptron (MLP) trained with dropout and L2 regularization, but without pre-training.  We configured an MLP network (MLP only architecture, see \autoref{table_archs}) with 50\% dropout on each layer and L2 regularization as in architecture D and suggested in \cite{srivastava2014dropout}. This architecture initially failed to converge over the first 200 epochs.  We then performed a sweep to characterize the parameter sensitivity of the MLP architecture. We found that the convergence of the model was highly sensitive to the learning rate; a change of \SI{1e-5} could cause the model to have no improvement over random chance. Choosing a learning rate of \SI{1.5e-5} as indicated by our parameter search, we then trained for the same number of epochs as our pre-trained architectures.  Although the initial convergence rate was similar, the MLP convergence became asymptotic at an error rate of 55\%. This asymptotic behavior was observed with stochastic gradient descent with momentum and with other learning rules such as Adaptive SubGradient (AdaGrad)~\cite{Duchi2011}. These results are in agreement with the work performed by ~\cite{Erhan2010};  they also indicate the challenge of using simple machine learning models to perform AMC.  Our results indicate that although it may be possible to configure an MLP such that it would converge for an AMC task, the relative robustness of the system is significantly reduced and the difficulty of parameter selection increases. By using unsupervised pre-training, we are able to substantially reduce parameter sensitivity and improve total training time and accuracy.  

\subsection{Regularization Assists Classification in Noisy Environments}
Regularization is typically prescribed in neural networks to prevent overfitting and to improve generalization. Unsupervised pre-training can also be considered a form of regularization, used to find a starting point such that final generalization error is reduced~\cite{Erhan2010}.  However, we have observed that, in an AMC task, regularization assists in classifying exemplars that are corrupted by effects \textit{not} found in the training set. We demonstrated this by examining the classification performance of our architectures against a dataset corrupted with additive white Gaussian noise (AWGN), a typical challenge in radio-frequency propagation testing. 

When classifying test samples from the test set which have been corrupted by noise, the most heavily regularized and pre-trained network tested (architecture D) exhibited the best overall performance. In the absence of noise, the best performance was observed in the unconstrained single-layer architecture ($P_{cc}=99.91\%$). To quantify performance in the presence of noise, we can examine the SNR required to achieve a performance of a specific $P_{cc}$, e.g. $P_{cc}=90 \%$, or classification error $1-P_{cc}=10\% $. By this measure, the unconstrained single-layer network (A, \figref{fig:snr}) had poorer performance, requiring an SNR of $\approx$\SI{5}{\dB} to reach $P_{cc}=90\%$.  The addition of a second layer with constraints (C, \figref{fig:snr}) results in a modest improvement of $\approx$ 2 dB. When sparse pre-training and L2 regularization are included as constraints (D, \figref{fig:snr}), the same performance can be achieved at an SNR of -1 dB; this represents an improvement of $\approx$\SI{6}{\dB} over the unconstrained single-layer network.  This corresponds to a 4-fold increase in maximum noise level for a given detection rate. 

The addition of sparsity appears to be crucial to this performance increase, and may be a result of forcing the selection of the most valuable receptive fields (rather than simply the ones that best fit the training data)~\cite{Vincent:2010:SDA:1756006.1953039}. As can be seen in \figref{fig:snr}, as we incrementally released forms of regularization, the performance against noisy data (SNR <5 dB) decreased.  This is a particularly useful aspect of this implementation of SSDAs; as propagation of digitally-transmitted radio signals through real environments presents a significant modeling challenge, the ability of our AMC method to compensate in a model-free way for such noise is highly desirable.  

The performance of the single-layer architecture also indicates that addition of such regularizations can have drawbacks that must be compensated for; without a second layer, a fully regularized single-layer network does not converge to adequately high performance levels (B, \autoref{table_archs}); however, it does generalize better than an MLP alone in the presence of noise.  This may be because it must rely on a limited selection of receptive fields, and with a small network and strong constraints, there may not be enough neurons active to adequately represent the necessary features for classification. These same primitive features, however,  may remain intact during signal corruption and thus allow higher low-SNR performance.  

We also tested a deeper architecture to see if additional layers would improve overall classification, or outweigh the regularization effects and reduce generalization for un-trained environmental noise.  In prior work on deep neural network architectures, it is typical to find that adding a layer improves performance by less than 1\%~\cite{Erhan2010}, and in noise-free conditions this agrees with our results (see \autoref{table_archs}).  Our deeper model consisted of architecture D with an additional two layers, subject to similar sparsity contraints (see architecture E, \autoref{table_archs}).  Interestingly, this model converged to high accuracy very quickly.  Thus, the addition of additional pre-trained layers resulted in a rapidly converging, highly accurate classifier.  Unfortunately, this configuration also performed substantially worse when exposed to signals in an AWGN channel.  We hypothesize that this may be a somewhat desirable form of overfitting; by adding additional layers, our classifier becomes highly tuned to the properties of the input set but somewhat inflexible. To improve generalization, one could explore the use of convolutional networks to provide strong regularization (in terms of a limited number of shared receptive fields) while using a deeper representation to achieve high accuracy.  It is possible such a network may achieve the rapid convergence seen with our deep SSDA, but without the loss in performance in the presence of un-modeled noise.

\subsection{Performance Per-Type}

Some insights come from studying how the classifier begins to fail under noisy conditions, as we show in ~\figref{fig:precision} and \figref{fig:confusion-matrix}. The confusion matrix shows the full distribution of $P_c^{(i|i')}$ at selected SNRs, and the precision and sensitivity curves show the full behavior of the marginal distribution. Recall that precision measures, within the set of samples predicted to have a given modulation, the fraction that actually have that modulation. Sensitivity measures, within the set of samples that actually have a given modulation, the fraction predicted to have that modulation. Precision for a class can be high if we correctly identify only a single example of that class; sensitivity for a class can be high if we assign every sample to that class. Our results show that the degradation in performance under noise is not random; for example, the classification system systematically over-predicts GMSK (as seen both in the corresponding columns of~\figref{fig:confusion-matrix} and the GMSK precision curve in~\figref{fig:precision}). Moreover, these degradations are not simply magnifications of the same errors that exist with no noise --- OFDM is the clearest example of this, as the system loses sensitivity to this modulation much more slowly than the other families. This behavior is likely an indicator of ``crosstalk'' in the receptive fields of the classification system. Where a traditional AMC architecture would rely on features that selected for a specific modulation family, our system learns features that are used for classifying multiple families: a single feature vector (receptive field) might play a role in reconstructing or identifying both GMSK and OFDM, for example, and the manner in which these vectors fail to fit noisy versions of their different target families is reflected in the way in which performance does not degrade uniformly for each family. A possible mitigation for this potential crosstalk may be as simple as adding more neurons to the autoencoder layers, as this will increase the number of possible receptive fields that our system learns. 

\subsection{Receptive Fields}
\begin{figure}
	\centering
	\includegraphics[trim=0.2in 0in  0.in 2.605in,clip, width=\linewidth, ]{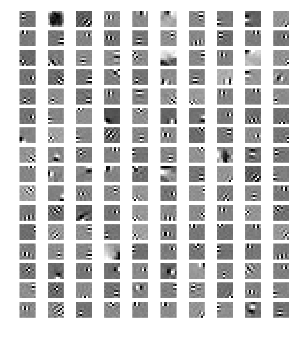}
	\caption{An example of Gabor-like layer 1 receptive fields generated by training an SSDA (architecture D, \autoref{table_archs}) on the Cifar-10 image dataset.}
	\label{fig:gabor}
\end{figure}

\begin{figure}
	\centering
	\includegraphics[trim=0.25in 0.75in  0.25in 0in,clip,width=3in]{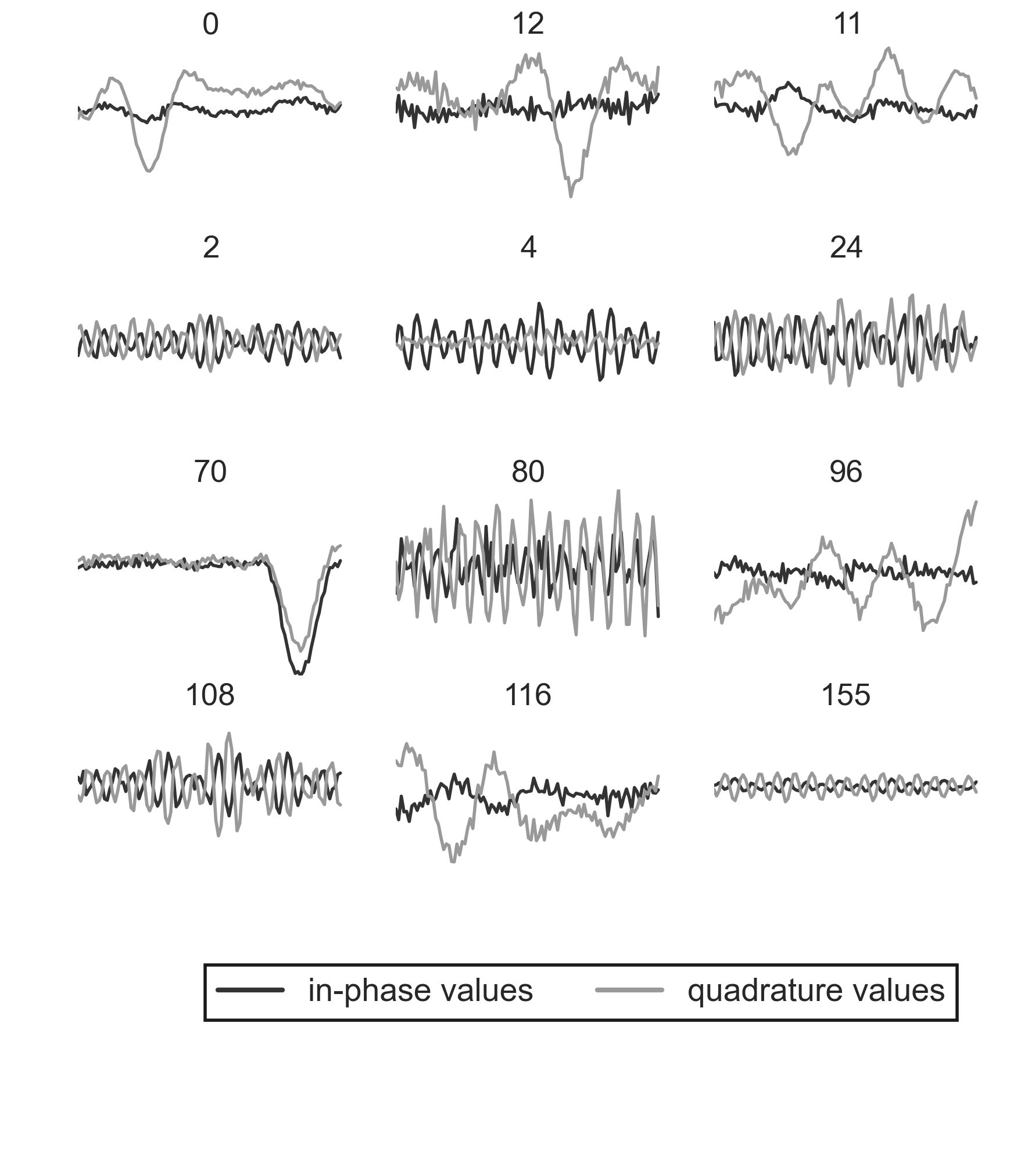}
	\caption{Examples of layer 1 receptive fields derived \autoref{table_archs} D. The number indicates the neuron to which the receptive field corresponds.  Each receptive field consists of the input that maximally activates a target neuron, and corresponds to the extracted features learned by the network.}
	\label{fig:rfs}
\end{figure}

The use of unsupervised feature extraction raises an important question: What sort of signal features is our system becoming sensitive to? The receptive fields in an autoencoder system are simply the weights between the input layer and the target layer, and they describe the input that maximally excites the target neuron. These features can be thought of as the primitive features of the input.   We began by verifying that our network would produce Gabor-like receptive when trained on the CIFAR-10 image dataset, which it did (see \figref{fig:gabor}). This aligns nicely with work presented in other papers, including work on dimensionality reduction in encoding~\cite{NIPS2011_1115}.  Although we did not analyze second-layer receptive fields, we chose to utilize sparsity in order to benefit from improvements in receptive fields observed by researchers studying receptive fields in sparse restricted Boltzmann machine models of visual cortex \cite{lee2008sparse}. In \figref{fig:data_modulations}, we present characteristic I/Q inputs from each modulation scheme; in \figref{fig:rfs} we present a selection of layer 1 receptive fields. The layer 1 I/Q receptive fields do not appear to exhibit strong alignment to an exact modulation type or symbol pattern; although that would have made for a neat explanation, it is perhaps not surprising.  While outside the scope of this paper, an analysis of the spectral and temporal characteristic of the autonomously determined receptive fields would provide a deeper understanding of what modulation features the system is learning.

The receptive fields shown in \figref{fig:rfs} represent the most significant difference between our research and prior studies with artificial neural network (ANN) based automatic modulation schemes. While other researchers (\cite{Azzouz2013, Wong2004}) have used single- and multi-layer ANNs to achieve impressive performance (as high as 97\% accuracy at 0 dB in a 10-way classification task~\cite{Wong2004}), these methods required expert construction of features specific to each modulation. Thus, these methods could be described as FB AMC with feature weighting. FB AMC with feature weighting is a significant contribution, although its reliance on detailed knowledge of signal characteristics makes it inflexible. Unsupervised methods such as those we present allow for much greater flexibility in terms of incorporating unusual characteristics of environmental noise, accommodating signals for which no detailed model may be available, and in adapting to changes in environmental or signal characteristics over time through the use of on-line learning techniques. ANN based methods are also actively being researched for use within the radio front-end processing stages. One example is to use a multilayer perceptron for channel equalization~\cite{Burse2010}. These efforts are orthogonal and potentially complementary to ours.

\subsection{Complementary Domains and Methods}
The task of automatic modulation classification for radio signal data is conceptually similar to tasks from related fields, such as phoneme or word detection in speech processing \cite{Deng2013, Hinton2012}, although the domain presents unique challenges in terms of sample rate and robustness to noisy environments. We also note that recent work in acoustic modeling with deep networks has found that significant improvements are possible by leveraging up to 7 layers of autoencoder units~\cite{Deng2013}, and the proof-of-concept architecture we present here will likely permit many more optimizations.  Additional improvements may come in the form of convolutional autoencoders~\cite{masci2011stacked}; as can be seen in \figref{fig:rfs} (receptive fields 0 and 12), some of the low-level features may be time-shifted variants of each other. This implies that convolutional application of those features to streaming inputs may provide performance and computational cost improvements.

Another possible route towards improved performance, especially in the application to streaming or online analysis, is the implementation of our architecture as a spiking neural network. Spiking neural networks (SNNs)~\cite{maass1997networks, paugam2012computing} are another step towards biologically-relevant systems, as they seek to represent information as discrete temporal events much like biological nervous systems do with action potentials. SNNs can natively represent information contained in signal timing with higher resolution than clocked systems of equivalent sophistication, and open up a much larger parameter space for encoding information. They provide new opportunities for unsupervised learning (spike-timing dependent plasticity ~\cite{bi1998synaptic}), optimization (spiking neuron models \cite{goodman2008brian, nageswaran2009configurable}), and efficient bandwidth usage (spike coding~\cite{thorpe2001spike}). We have implemented architecture D (\autoref{table_archs}) as an SNN, and have demonstrated near-identical performance on the same task described here in full spiking simulation. Our SNN results will be the focus of a companion paper to this work.  

\section{Conclusion}
\label{conclusion}

We have presented a fundamentally different way of addressing the challenge of automatic modulation classification (AMC) in the radio frequency domain. Our method of biologically-inspired feature extraction, feature abstraction (in which we construct more complex features), and labeling demonstrate that principles of animal sensory systems may be applied to achieve useful performance in an AMC task.  However, there is nothing fundamentally important about using I/Q samples, or even 2-dimensional input vectors. Our method is essentially a sensor-agnostic sensory system for providing classification in noisy environments.   

It is important to note that we did not perform any exhaustive hyperparameter searches or optimizations.  Rather than searching for an optimal network for this task (and we make no claims that our network is optimal), we approached the task of modulation classification from a biologically-inspired perspective.  In computer vision research, the realization that experimentally measured V1 receptive fields are similar to Gabor filters ~\cite{pmid14449617, pmid3437332} and the discovery that sparse coding of natural images generates similar fields ~\cite{Olshausen1996} is a crucial result.  Additional  work on recovering biologically-relevant receptive fields for different sensory modalities and with different dimensionality-reduction techniques was also a crucial step towards the research we present~\cite{NIPS2011_1115}. We configured our network such that, given natural scene data, it would produce well-known Gabor-like receptive fields. Our prediction, which has been verified by our experiments, was that a system that produces biologically-relevant receptive fields for a biologically-relevant task would also produce useful receptive fields for a non-biological task. This is the extent of optimization used, and it results in a network capable of useful performance for automatic modulation classification.

Our results differ from much prior work in neural-network processing of time-varying signals (speech recognition, for example) by focusing narrowly on ingesting raw waveform data, rather than spectrogram or filter bank features, and extracting useful features for later tasks. We have thus taken a first step down a road called out in~\cite{Deng2013}, and have demonstrated that even relatively simple networks can do useful processing of radio signals with extremely limited samples and in the presence of environmental noise. Our results also differ from the prior work in AMC, as they do not make use of expert knowledge and can construct effective features that adapt to both signals and the propagation environment with competitive performance.  This opens up new opportunities for efficient use of an increasingly complex electromagnetic signaling environment. We also hope that our research will lead to application of additional biological inspiration to this problem both on the sensing end (such as RF cochleas \cite{mandal2006circuits}) and on the processing end, where we are currently researching spiking neural networks for creating more complete perception of the electromagnetic environment. 

We also demonstrate that biologically-inspired feature extraction, in the form of sparsity and unsupervised pre-training, can enhance neural-network AMC even under noise conditions not modeled in the training data. We demonstrate that as the persistence and level of the sparsity constraints increase, the general performance of our classifier improves in environmental conditions under which the classifier was not specifically trained. Under no noise, all explored architectures that successfully converge perform similarly well, but we find that biologically motivated principles result in a system which performs markedly better under environmental noise. This is particularly interesting in light of the prevailing explanations for the sparse coding principle~\cite{Bell1997, Olshausen1996}, among them robustness to environmental noise. Our results indicate that this principle is still valid and useful in problem domains that are rarely associated with sensing by natural organisms. We believe that the most important and broadly applicable conclusion of our work is that biologically-inspired sensing principles, implemented using hierarchical neural networks, do not require a biologically-inspired input. This suggests that other areas for which both machine and human perception are limited (network traffic, equipment temperature, power grids) may benefit from application of the methods we propose.

\ifCLASSOPTIONcaptionsoff
  \newpage
\fi



\printbibliography

\vfill\eject
\begin{IEEEbiographynophoto}{Benjamin Migliori}
Ben Migliori received his M.S. in Physics and his Ph.D. in physics and biophysics from the University of California, San Diego, in 2009 and 2013 respectively. He is currently a research scientist at SPAWAR Systems Center Pacific, the primary research arm of the Space and Naval Warfare Systems Command.  His research focuses on application of biological inspiration to scientific and technical challenges.
\end{IEEEbiographynophoto}
\begin{IEEEbiographynophoto}{Riley Zeller-Townsen}
	Riley Zeller-Townsen received his B.S. in Biomedical Engineering and Computer Engineering from North Carolina State University in 2009.  He is currently a graduate student at Georgia Institute of Technology and a Dept. of Defense SMART scholar.  His research interests include neural interfaces, biomimetic systems, and working towards conceptual frameworks that make sense of the brain. 
\end{IEEEbiographynophoto}
\begin{IEEEbiographynophoto}{Daniel Grady}
	Daniel Grady received his M.S. and Ph.D. in Engineering Science and Applied Mathematics from Northwester University in 2007 and 2012 respectively. His research interests focus on complex networks and graph models.  He is currently a data scientist at ID Analytics in San Diego. 
\end{IEEEbiographynophoto}
\begin{IEEEbiographynophoto}{Daniel Gebhardt}
Daniel Gebhardt received his B.S in Electrical Engineering from the University of Portland in 2004 and his Ph.D. in Computer Science from the University of Utah in 2011.  He is currently a research scientist at SPAWAR Systems Center Pacific, the primary research arm of the Space and Naval Warfare Systems Command.  His research interests include the development of low size, weight and power implementations of modern machine learning algorithms. 
\end{IEEEbiographynophoto}
\vfill




\end{document}